\pdfoutput=1

\documentclass[11pt]{article}

\usepackage{acl}
\usepackage{times}
\usepackage{latexsym}
\usepackage[T1]{fontenc}
\usepackage[utf8]{inputenc}
\usepackage{microtype}
\usepackage{scrextend}

\usepackage{graphicx}
\usepackage{xspace}
\usepackage{booktabs}
\usepackage{multirow, array}
\usepackage{amsmath}
\usepackage{amssymb}
\usepackage{amsfonts}
\usepackage{xcolor}
\usepackage{ulem}
\usepackage{cleveref}
\usepackage[ruled,vlined]{algorithm2e}
\SetAlFnt{\small}
\SetAlCapNameFnt{\small}
\SetAlCapFnt{\small}
\usepackage{enumitem}

\newcommand\hlt[1]{\textcolor{red}{\emph{#1}}}

\newcommand{\bx}{\mathbf{x}}

\newcommand{\bu}{\mathbf{u}}

\newcommand{\ppl}{\text{ppl}}

\newcommand{\model}{R\&R\xspace}

\begin{document}

\title{R\&R: Metric-guided Adversarial Sentence Generation}

\newcommand*{\affaddr}[1]{#1} 
\newcommand*{\affmark}[1][*]{\textsuperscript{#1}}
\newcommand*{\email}[1]{\texttt{#1}}

\author{
Lei Xu\affmark[1], Alfredo Cuesta-Infante\affmark[2], Laure Berti-Equille\affmark[3], Kalyan Veeramachaneni\affmark[1] \\
\affaddr{\affmark[1] MIT LIDS}  \hspace{1ex}
\affaddr{\affmark[2] Universidad Rey Juan Carlos}  \hspace{1ex}
\affaddr{\affmark[3] IRD}\\
\email{leix@mit.edu \hspace{1ex} alfredo.cuesta@urjc.es \hspace{1ex} laure.berti@ird.fr} \\ 
\email{kalyanv@mit.edu}
}
\maketitle

\begin{abstract}
Adversarial examples are helpful for analyzing and improving the robustness of text classifiers.
Generating high-quality adversarial examples is a challenging task as it requires generating fluent adversarial sentences that are semantically similar to the original sentences and preserve the original labels, while causing the classifier to misclassify them.
Existing methods prioritize misclassification by maximizing each perturbation's effectiveness at misleading a text classifier; thus, the generated adversarial examples fall short in terms of fluency and similarity. 
In this paper, we propose a rewrite and rollback (\model) framework for adversarial attack. It improves the quality of adversarial examples by optimizing a critique score which combines the fluency, similarity, and misclassification metrics. 
\model generates high-quality adversarial examples by allowing exploration of perturbations that do not have immediate impact on the misclassification metric but can improve fluency and similarity metrics.
We evaluate our method on 5 representative datasets and 3 classifier architectures. Our method outperforms current state-of-the-art in attack success rate by +16.2\%, +12.8\%, and +14.0\% on the classifiers respectively. 
Code is available at \url{https://github.com/DAI-Lab/fibber}
\end{abstract}

\section{Introduction}

Recently, adversarial attacks in text classification have received a great deal of attention. Adversarial attacks are defined as subtle perturbations in the input text such that a classifier misclassifies it.
They can serve as a tool to analyze and improve the robustness of text classifiers, thus being more and more important because security-critical classifiers are being widely deployed~\cite{wu2019misinformation, torabi2019big, zhou2019fake}.

\begin{figure}[htb]
    \centering
    \includegraphics[width=\columnwidth]{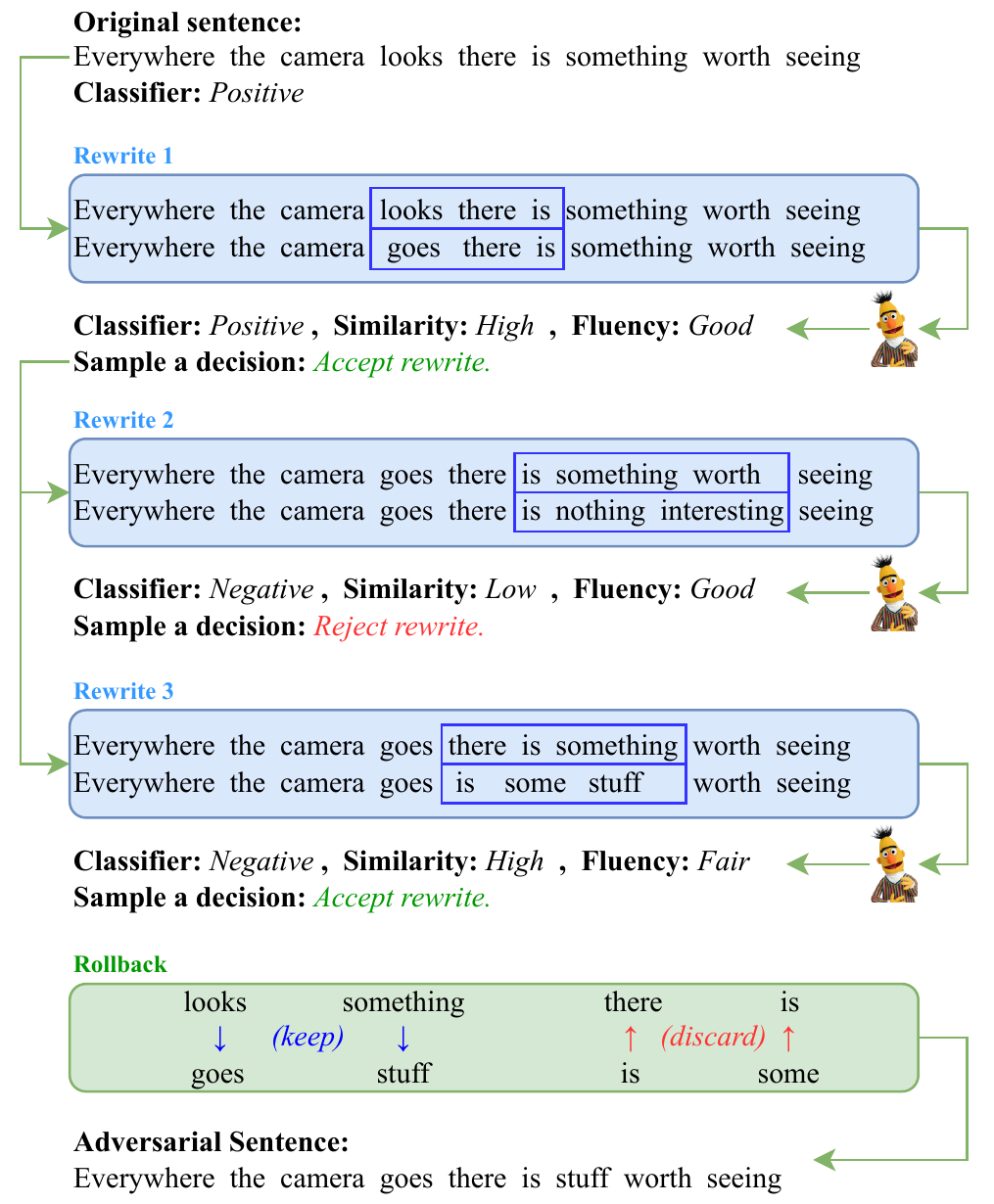}
    \caption{\model generates adversarial examples by rewrite and rollback. The rewrite step  explores possible perturbations stochastically and is guided by similarity metric and fluency metric to ensure better quality of the example. The rollback operation further improves the similarity.}
    \label{fig:eg}
    \vspace{-1em}
\end{figure}

Existing attack methods either adopt a synonym substitution approach~\citep{Jin2019IsBR, zang2019word} or use a pre-trained language model to propose substitutions for better fluency and naturalness~\citep{li2020bert, garg2020bae, li2021CLARE}. They follow a similar framework: first, construct some candidate perturbations, and then, use the perturbations that most effectively mislead the classifier to modify the sentence. This process is repeated multiple times until an adversarial example is found. This framework prioritizes misclassification by picking perturbations that most effectively mislead the classifier. Despite the success in changing the classifier prediction, it has two main disadvantages. 
First, it is prone to modify words that are critical to the sentence's meaning which decreases the similarity and is more likely to change the true label of the sentence, or introduce low-frequency words causing the fluency to decrease.
Second, some perturbations do not have immediate impacts on misclassification, but can trigger it when combined with other perturbations, and these frameworks cannot find adversarial examples with these perturbations.

To overcome these problems, the attack method needs to consider fluency, similarity, and misclassification jointly, while also efficiently exploring various perturbations that do not show direct impacts on the latter. We define a critique score that combines fluency, similarity and misclassification metrics. Then, we present our design for a Rewrite and Rollback framework~(\model) which optimizes this score to generate better adversarial examples. 
In the rewrite stage, we explore multi-word substitutions proposed by a pre-trained language model. We accept or reject a substitution according to the critique score. We can generate a high-quality adversarial example after multiple iterations of rewrite.
Rewrite may introduce changes that do not contribute to misclassification and may also reduce similarity and fluency. Therefore, we periodically apply the rollback operation to reduce the number of modifications without changing the misclassification result. 
Figure~\ref{fig:eg} illustrates the process using an example.



\begin{figure*}[htb]
    \centering
    \includegraphics[width=\textwidth]{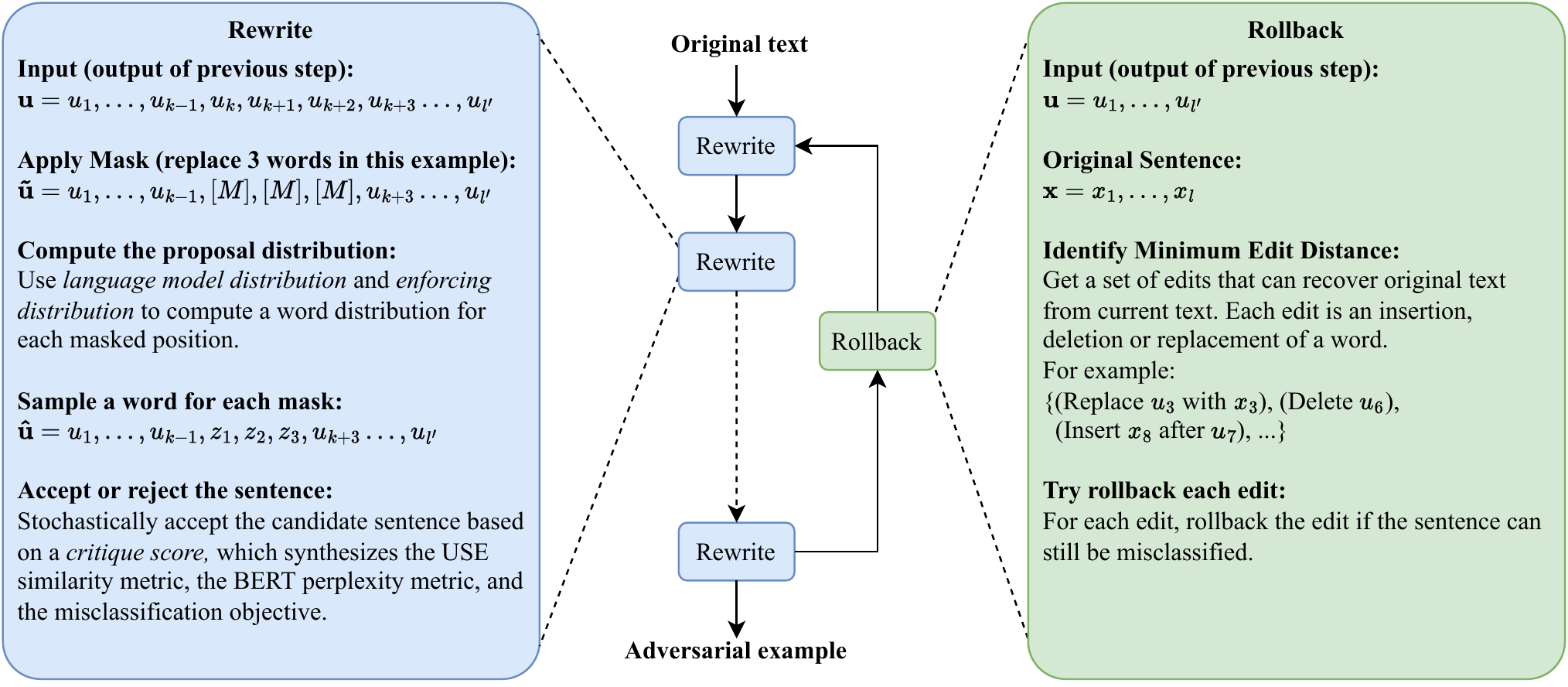}
    \caption{\model Framework.}
    \label{fig:framework}
\end{figure*}

\section{Problem Formulation}
Let $\bx=x_1,\ldots, x_l$ be a sentence of length $l$, $y$ be its classification label, and $f(\bx)$ be a text classifier that predicts a probability distribution over classes. The objective of an attack method $\mathcal{A}(\bx, y, f)$ is to
\begin{align*}
& \text{construct } \bu=u_1,\ldots, u_{l'} \text{ satisfying 3 conditions:} \\
&\begin{cases}
     \text{\small $\bu$ is misclassified, i.e., $f(\bu)\neq y$,}\\
     \text{\small Human considers $\bu$ as a fluent sentence,}\\
     \text{\small Human considers $\bu$ to be semantically similar to $\bx$.}\\
     \text{\small Human considers $\bu$ preserves the true label $y$.}
\end{cases}
\end{align*}
  where $l'$ is the length of the adversarial sentence. 
However, this formulation requiring human evaluation is intractable for large-scale data. 
Therefore, we approximate the sentence fluency with the perplexity of the sentence. It is defined as 
\[
\textstyle
\ppl(\bx) = \exp\big[-\frac{1}{l}\sum_{i=1}^{l}\log p(x_i|x_1\ldots x_{i-1})\big],
\]
where $p(x_i|x_1\ldots x_{i-1})$ is measured by a language model. Low perplexity means the sentence is predictable by the language model, which usually indicates the sentence is fluent. 
Sentence similarity can be quantified as $\cos\big(H(\bx), H(\bu)\big)$,
where $H(\cdot)$ is a pre-trained sentence encoder that encodes the meaning of a sentence into a vector. 
We assume that high sentence similarity implies preservation of the sentence label. 
%
Thus, finding the adversarial sentence $\bu$ is formulated as a multi-objective optimization problem as follows:
\begin{align*}
     \text{Construct } & \bu=u_1,\ldots, u_{l'}
\text{  to } \text{minimize } \ppl(\bu) \\
              &\text{ and }\text{maximize } \cos\big(H(\bx), H(\bu)\big)\\
&\text{  subject to } f(\bu)\neq y.
\end{align*}
We use a fine-tuned BERT-base model~\citep{Devlin2019BERTPO} to measure perplexity and use Unversal Sentence Encoder (USE)~\citep{cer2018universal} to measure sentence similarity. Ultimately, fluency, similarity, and the preservation of original label need to be verified by humans. We discuss human verification in Section~\ref{sec:exp}.

\paragraph{Threat Model. } We assume the attacker can query the classifier for the prediction (i.e., the probability distribution over all classes). But they do not have knowledge on architecture of the classifier nor query for the gradient. They can also access some unlabeled text in the domain of the classifier.

\section{Metric-Guided Rewrite and Rollback}
In this section, we first give an overview, then introduce the rewrite and rollback components respectively. Finally, we give a summary of pre-trained models used in the framework.

\subsection{Overview}

\model contains the rewrite and rollback steps. 
In the rewrite step, we randomly mask several consecutive words, and compute a \textit{proposal distribution}, which is a distribution over the vocabulary on each masked position defined as Eq.~(\ref{eq:proposal}). We construct a multi-word substitution\footnote{The number of words in each substitution, the number of rewrite steps between two rollback steps,  the maximum number of rewrite steps, and the batch size are hyperparamters.\label{ishp}} for the masked positions according to the distribution, then compute the \textit{critique score} defined as Eq.~(\ref{eq:pplcc})-(\ref{eq:clfcc}). 
If the score increases, we accept the substitution. If the score decreases, we accept it with a probability depending on the degree of decrease. The rewrite step contains randomness to encourage exploration of different modifications, while the critique score will guide the rewritten sentence to a high-quality adversarial example.
After several steps of rewriting\footref{ishp}, we apply a rollback operation on the sentences that have already been misclassified to reduce the number of changes introduced in the rewriting. In the rollback step, we identify a minimum set of edits required to change the current sentence back to the original sentence. We rollback an edit if it does not affect the misclassification. 

We implement the framework to simultaneously rewrite a batch of sentences. We make multiple copies of an input text and create a batch\footref{ishp}. The proposal distributions and critique scores for these copies can be computed in parallel on a GPU, while the randomness in the rewrite step leads to different rewritten sentences. 
The loop terminates when either the maximum number of rewrite steps is reached\footref{ishp} or half of the sentences in the batch are misclassified. 
Figure~\ref{fig:framework} shows the \model framework.

\subsection{Rewrite}
In each rewrite, we mask then substitute a span of words. It is composed of the following steps.

\paragraph{Apply mask in the sentence. }
First, we randomly pick $m$ consecutive words in the sentence, and replace them with $t$ mask, where $t$ can be $m$, $m-1$, or $m+1$ meaning \textit{replace}, \textit{shrink}, and \textit{expand} operation respectively. Compared with CLARE~\cite{li2021CLARE} which masks one word at a time (i.e., $m=1$), masking multiple words can make it easier to modify common phrases. We use $\tilde\bu$ to denote the masked sentence. 

\paragraph{Compute proposal distribution. }
Then, we compute proposal distribution for $t$ masks in the sentence. This distribution assigns a high probability to words that can construct a fluent and legitimate paraphrase. Let $z_1,\ldots z_t$ be the words to be placed at the masked positions. The distribution is 
\begin{equation}
\textstyle
p_\text{proposal}(z_i|\tilde\bu, \bx) \propto  p_\text{lm}(z_i|\tilde\bu)  \times p_\text{enforce}(z_i|\tilde\bu, \bx)\label{eq:proposal}
\end{equation}
where $p_\text{lm}$ is a \textit{language model distribution} that give high probability to words that can make a fluent sentence, and $p_\text{enforce}$ is the \textit{enforcing distribution}, which give high probability to words that can lead to semantically similar sentences. $p_\text{lm}$ and $p_\text{enforce}$ should be considered as two different weights of words and are multiplied together to get $p_\text{proposal}$ so that if either $p_\text{lm}$ or $p_\text{enforce}$ is low, the word will have low probability in $p_\text{proposal}$. This is a desired property because we want the adversarial sentence to have good fluency (i.e., high $p_\text{lm}$) and high similarity (i.e., high $p_\text{enforce}$).
%
$p_\text{lm}$ is computed by sending $\tilde \bu$ into BERT and taking the predicted word distribution on masked positions. Depending on the position, the word distributions for $t$ masks are different.
The enforcing distribution is measured by word embeddings. We use the sum of word embeddings $R(\bu)=\sum_{u_k} E(u_k)$ as a sentence embedding, where $E(\cdot)$ is the counter-fitted word embedding~\citep{mrkvsic2016counter}. Then we define the enforcing distribution as 
\begin{align}
p_{\text{enforce}}&(z_i|\tilde\bu, \bx)\propto \exp\big[w_{\text{enforce}}\nonumber\\
&\times (\cos(R(\bx)-R(\tilde\bu), E(z_i)) - 1)\big].\label{eq:enforce}
\end{align}
$w_{\text{enforce}}$ is a hyper-parameter with a positive value. Larger $w_{\text{enforce}}$ penalizes more on dissimilar words. The $\exp$ ensures the value to be positive thus the values can be converted to a probability distribution over words. We use the conventional cosine similarity to compute the distance of two vectors. If the embedding of a word $E(z)$ perfectly aligns with the sentence representation difference $R(\bx)-R(\tilde\bu)$, it gets the largest probability. The enforcing distribution aims at making the candidate modification more similar to the original sentence. Note that enforcing distribution is identical on all $t$ masks.

\paragraph{Sample a candidate sentence.} 
We sample a candidate word for each masked position by $z_i\sim p_{\text{proposal}}(z_i|\tilde\bu, \bx)$. We do not consider the effect of sampling one word on other masked positions (i.e., we do not recompute proposal distribution for the remaining masks after sampling a word) because language model distribution already considers the position of the mask and assigns a different distribution for each mask, meanwhile recomputing is inefficient. We use $\hat\bu$ to denote the candidate sentence. 

\paragraph{Critique score and decision function.} 
We decide whether to accept the candidate sentence using a decision function. The decision function computes a heuristic critique score
\begin{align}
\textstyle
C&(\bu) = \big(w_\text{ppl} \min(1-\ppl(\bu)/\ppl(\bx), 0) \label{eq:pplcc}\\
&+w_\text{sim} \min(\cos\big(H(\bu), H(\bx)\big) - \phi_\text{sim}, 0)\label{eq:simcc}\\
&+w_\text{clf} \min(\max_{y'\neq y}f(\bu)_{y'} - f(\bu)_y, 0)\big)
\label{eq:clfcc}
\end{align}
Eq.~(\ref{eq:pplcc}) penalizes sentences with high perplexity, where $\ppl(\bx)$ is perplexity measured by a BERT model. Eq.~(\ref{eq:simcc}) penalizes sentences with sentences with cosine similarity lower than $\phi_\text{sim}$, where $H(\cdot)$ is the sentence representation by USE.
Eq.~(\ref{eq:clfcc}) penalizes sentences that cannot be misclassified where $f(\bu)_y$ means the log probability of class $y$ predicted by the classifier. $w_\text{ppl}$, $w_\text{sim}$ and $w_\text{clf}$ are hyperparameters.

The decision is made based on 
\begin{equation}
\textstyle
    \alpha=\exp[C(\hat\bu) - C(\bu)]. \label{eq:alpha}
\end{equation} 
If $\alpha > 1$, the decision function accepts $\hat\bu$; otherwise it accepts $\hat\bu$ with probability $\alpha$.
The computation of $\alpha$ is motivated by the Metropolis–Hastings algorithm~\citep{hastings1970monte} (See Appendix~\ref{app:mhs}).
The critique score is a straightforward way to convert the multi-objective optimization problem into a single objective. Although it introduces several hyper-parameters, \model is no more complicated than conventional methods, which also require hyper-parameter setting. 

\subsection{Rollback}
In the rollback step, we eliminate modifications that do not correct the misclassification. It contains the following steps. 

\paragraph{Find a minimum set of simple edits.} 
We first find a set of simple edits that change the current rewritten sentence back to the original sentence. Simple edits mean the insertion, deletion or replacement of a single word, which is different from the modification in the rewrite step. 

\paragraph{Rollback edits.} 
For each edit, if reverting it does not correct the misclassification, then we revert the edit. For convenience, we scan each word in the sentence from right to left, and try to rollback each edit. Note that rollback may introduce grammar errors, but they can be fixed in future rewrite steps. 

\subsection{Vocabulary Adaptation} 
Computing $p_\text{propose}$ is challenging because of the inconsistent vocabulary. The counter fitted word embeddings in $p_\text{enforce}(\cdot)$ works on a 65k-word vocabulary, while the BERT language model used in $p_\text{lm}(\cdot)$ uses a 30k-word-piece vocabulary which contains common words and affixes. Rare words are handled as multiple affixes. For example ``hyperparameter'' does not appear in the BERT vocabulary, so it is handled as ``hyper'', ``\#\#para'', and ``\#\#meter''. 
Since the BERT model is more complicated, we keep it as is and transfer word embeddings to BERT vocabulary. 
We train the word-piece embeddings as follows. Let $\mathbf{w}=\{w_1, \ldots, w_L\}$ be a plain text corpus tokenized by words. Let $T(w)$ be word-piece tokenization of a word. Let $E(w)$ be the original word embeddings and $E'(x)$ be the transferred embeddings on word-piece. 
We train the word-piece embeddings $E'$ by minimizing the absolute error $\sum_{w\in\mathbf{w}}||E(w)-\sum_{x\in T(w)}E'(x)||_1$.
We initialize $E'$ by copying the embedding on words shared by two vocabularies and set other embeddings to $0$. We optimize the absolute error using stochastic gradient descent. In each step, we sample 5000 words from $\mathbf{w}$, then update $E'$ accordingly. Figure~
\ref{fig:adapt} in Appendix illustrates the algorithm.

\subsection{Summary of pre-trained models in \model}
In \model, we employ several pre-trained models. Choices are made according to the different characteristics of these pre-trained models.

\noindent\textbf{BERT for masked word prediction and perplexity.} Because BERT is originally trained for masked word prediction, it can predict the word distribution given context from both sides. Thus, BERT is preferable for generating $p_\text{lm}$. Estimating the perplexity for a sentence requires BERT to run in decoder mode and be fine-tuned. Perplexity can also be measured by other language models such as GPT2~\cite{radford2019language}. We use BERT mainly for the consistent vocabulary with $p_\text{lm}$.

\noindent\textbf{Word embedding and USE for similarity.}  Word embedding is more efficient as it only computes the sum of vectors and cosine similarity. In enforcing distribution, we need to replace the selected position with all possible $z$'s and measure the similarity, so we use word embeddings for efficiency. In the critique score, only the proposal sentence needs to be measured, so we can afford more computation time of USE.

\section{Experiments}\label{sec:exp}
We conducted experiments on a wide range of datasets and multiple victim classifiers to show the efficacy of \model. We first evaluate the quality of adversarial examples using automatic metrics. Then, we conducted human evaluation to show the necessity to generate highly similar and fluent adversarial examples. Finally, we conduct an ablation study to analyze each component of our method, and discuss defense against the attack.

\noindent\textbf{Datasets.}  We use 3 conventional text classification datasets: topic classification, sentiment classification, and question type classification. We also use 2 security-critical datasets: hate speech detection and fake news detection. Dataset details are given in Table~\ref{tab:dataset}.


\begin{table}[htb]
\centering\small
\begin{tabular}{cccp{4cm}}
\toprule
\textbf{Name} & \textbf{\#C} & \textbf{Len} & \textbf{Description}\\
\midrule
AG  & 4 & 43 & News topic classification by \citet{zhang2015character}.    \\
MR & 2 & 32 & Moview review dataset by \citet{pang2005seeing}.\\
TREC & 6 & 8 & Question type classification by \citet{li2002learning}. \\
HATE & 2 & 23 & Hate speech detection dataset by \citet{kurita20acl}.\\
FAKE & 2 & 30 & Fake news detection dataset by \citet{yang2017satirical}. We use the first sentence of the news for classification.\\
\bottomrule
\end{tabular}
\caption{Dataset details. \#C means number of classes. Len is the average number of words in a sentence. }
\label{tab:dataset}
\end{table}

\begin{table}[htb]
\small\centering
\begingroup
\setlength{\tabcolsep}{4pt}
\begin{tabular}{rccccc}
\toprule
              & \textbf{AG}   & \textbf{MR}   & \textbf{TREC} & \textbf{HATE} & \textbf{FAKE} \\\midrule
BERT-base     & 92.8 & 88.2 & 97.8 & 94.0 & 81.2      \\
RoBERTa-large & 92.7 & 91.6 & 97.3 & 95.0 & 75.5      \\
FastText      & 89.2 & 79.5 & 85.8 & 91.5 & 72.4      \\\midrule
Log Perplexity    & 3.38 & 5.27 & 3.91 & 3.56 & 4.92      \\\bottomrule
\end{tabular}
\endgroup
\caption{Accuracy of 3 classifers and sentence log perplexity on the clean test set.}
\end{table}
\begin{table*}[htb]
\centering\small
\begingroup
\setlength{\tabcolsep}{4pt}
\begin{tabular}{lrccccccccccccccc}
\toprule
 &             & \multicolumn{3}{c}{\textbf{AG}} & \multicolumn{3}{c}{\textbf{MR}} & \multicolumn{3}{c}{\textbf{TREC}} & \multicolumn{3}{c}{\textbf{HATE}} & \multicolumn{3}{c}{\textbf{FAKE}} \\\cmidrule(lr){3-5}\cmidrule(lr){6-8}\cmidrule(lr){9-11}\cmidrule(lr){12-14}\cmidrule(lr){15-17}
 & \textbf{Attack}    & \textbf{ASR}  & \textbf{Sim}  & \textbf{PPL}  & \textbf{ASR}  & \textbf{Sim}  & \textbf{PPL}  & \textbf{ASR}  & \textbf{Sim}   & \textbf{PPL}   & \textbf{ASR}  & \textbf{Sim}   & \textbf{PPL}   & \textbf{ASR}    & \textbf{Sim}     & \textbf{PPL}    \\\midrule
\multirow{3}{*}{\rotatebox[origin=c]{90}{\scriptsize BERT}}
 & TextFooler  & 16.8   & \textbf{0.98}  & 4.00  & 26.0   & 0.97  & 5.92  & 1.8    & \textbf{0.97}   & 5.30   & 30.6   & 0.97   & \textbf{3.53}   & 29.9     & \textbf{0.98}     & 5.44    \\
 & CLARE       & 28.8   & 0.97  & \textbf{3.60}  & 48.4   & 0.97  & 5.70  & 2.5    & 0.96   & 5.58   & 31.0   & 0.97   & 3.99   & 48.9     & \textbf{0.98}     & \textbf{5.02}    \\
&\model (Ours) & \textbf{54.1}   & \textbf{0.98}  & 3.64  & \textbf{63.4}   & \textbf{0.98}  & \textbf{5.36}  & \textbf{10.8}   & \textbf{0.97}   & \textbf{5.29}   & \textbf{55.3}   & \textbf{0.98}   & 4.06   & \textbf{57.0}     & \textbf{0.98}     & 5.05   \\\midrule

\multirow{3}{*}{\rotatebox[origin=c]{90}{\scriptsize RoBERTa}}
&TextFooler   & 15.6 & \textbf{0.98} & 5.21 & 18.0 & \textbf{0.97} & 6.06 & 0.4  & 0.96 & 7.09 & 24.0 & \textbf{0.98} & 4.20  & 26.6 & \textbf{0.98} & 5.45 \\
&CLARE        & 23.3 & 0.97 & 5.24 & 45.9 & \textbf{0.97} & 5.67 & 2.5  & \textbf{0.97} & 6.53 & 35.7 & 0.97 & 4.37 & 46.0 & \textbf{0.98} & \textbf{5.20} \\
&\model (Ours)& \textbf{41.2} & \textbf{0.98} & \textbf{3.73} & \textbf{48.5} & \textbf{0.97} & \textbf{5.53} & \textbf{12.5} & \textbf{0.97} & \textbf{5.17} & \textbf{55.7} & 0.97 & \textbf{4.07} & \textbf{59.6} & \textbf{0.98} & 5.25 \\\midrule
 
\multirow{3}{*}{\rotatebox[origin=c]{90}{\scriptsize FastText}}
& TextFooler  & 25.8 & \textbf{0.98} & 4.16 & 33.1 & \textbf{0.98} & 5.85 & 6.5  & \textbf{0.98} & 5.04 & 21.7 & \textbf{0.98} & \textbf{3.44} & 35.3 & \textbf{0.98} & 5.46 \\
& CLARE       & 28.9 & 0.97 & 3.91 & 41.5 & 0.97 & 5.79 & 8.5  & 0.97 & 6.06 & 35.6 & 0.97 & 4.24 & 76.0 & \textbf{0.98} & 5.15 \\
&\model (Ours)& \textbf{37.8} & \textbf{0.98} & \textbf{3.84} & \textbf{48.9} & \textbf{0.98} & \textbf{5.48} & \textbf{44.1} & \textbf{0.98} & \textbf{4.68} & \textbf{53.3} & \textbf{0.98} & 4.03 & \textbf{76.4} & \textbf{0.98} & \textbf{5.10} \\\bottomrule
\end{tabular}
\caption{Automatic evaluation results. ``Sim'' and ``PPL'' represent similarity measured by USE and the log perplexity measured by BERT respectively.}\label{tab:auto_metrics}
\endgroup
\end{table*}


\noindent\textbf{Victim Classifiers.} For each dataset, we use the full training set to train three victim classifiers:  
   (1) BERT-base classifier~\cite{Devlin2019BERTPO}; 
    (2) RoBERTa-large classifier~\cite{liu2019roberta}, and 
    (3) FastText classifier~\cite{joulin2017fasttext}.

\noindent\textbf{Baselines.} We compare our method against two strong baseline attack methods: 
    TextFooler~\citep{Jin2019IsBR} and 
    CLARE~\citep{li2021CLARE}.

\noindent\textbf{Hyperparameters.} In \model, we use the BERT-base language model for $p_\text{lm}$. For each dataset, we fine-tune the BERT language model using 5k batches on the training set\footnote{We use the plain text to fine-tune the language model, and do not use the label. In the threat model, we assume the attacker can access plain text data from a similar domain. } with batch size 32 and learning rate 0.0001, so it is adapted to the dataset. We set the enforcing distribution hyper-parameters $w_{\text{enforce}}=5$. The decision function hyper-parameters $w_{\text{ppl}} = 5$, $w_{\text{sim}}=20$, $\phi_{\text{sim}}=0.95$, $w_{\text{clf}}=2$. To generate each paraphrase, we set maximum rewrite iterations to be $200$, and replace a 3-word span in each iteration. We implement \model in a 50-sentence batch and apply early-stop when half of the batch are misclassified. We apply rollback operation every 10 steps of rewrite. Then, we return the adversarial example with the best critique score. 

\noindent\textbf{Hardware and Efficiency.} We conduct experiments on Nvidia RTX Titan GPUs. We measure the efficiency using average wall clock time. On the MR dataset, one attack on a BERT-base classifier using \model takes 15.8 seconds on average. CLARE takes 14.4 seconds on average. TextFooler is the most efficient algorithm which takes 0.45 seconds. 

\noindent\textbf{Automatic Metrics.} We evaluate the efficacy of the attack method using 3 automatic metrics:

\noindent \underline{\textit{Similarity ($\uparrow$)}:} We use Universal Sentence Encoder to encode the original and adversarial sentence, then use the cosine distance of two vectors to measure the similarity. We set a similarity threshold at 0.95, so the similarity of a legitimate adversarial example should be greater than 0.95.

\noindent \underline{\textit{Log Perplexity ($\downarrow$)}} shows the fluency of adversarial sentences. 

\noindent \underline{\textit{Attack success rate (ASR) ($\uparrow$)}} shows the ratio of correctly classified text that can be successfully attacked.

\noindent\textbf{Human Metrics}: Automatic metrics are not always reliable. We use Mechanical Turk to verify the similarity, fluency, and whether the label of the text is preserved with respect to human evaluation. 

\noindent \underline{\textit{Sentence similarity} ($\uparrow$):} Turkers are shown pairs of original and adversarial sentences, and are asked whether the two sentences have the same semantic meaning. They annotate the sentence in a 5-likert, where 1 means strongly disagree, 2 means disagree, 3 means not sure, 4 means agree, and 5 means strongly agree. 

\noindent \underline{\textit{Sentence fluency} ($\uparrow$):} Turkers are shown a random shuffle of adversarial sentences, and are asked to rate the fluency in a 5-likert, where 1 describes a bad sentence, 3 describes a meaningful sentence with a few grammar errors, and 5 describes a perfect sentence.

\noindent \underline{\textit{Label match} ($\uparrow$):} Turkers are shown a random shuffle of adversarial sentences and are asked whether it belongs to the class of the original sentence. They are asked to rate 0 as disagree, 0.5 as not sure, and 1 as agree. 

We sample 100 adversarial sentences from each method, and each task is annotated by 2 Turkers. We do not annotate label matches on the FAKE dataset because identifying fake news is too challenging for Turkers. We require the location of the Turkers to be in the United States, and their Hit Approval Rate to be greater than 95\%. The screenshots of the annotation tasks are shown on Figure~\ref{fig:mturk_screen} in Appendix.


\noindent\textbf{Examples.} Table~\ref{tab:case} shows some examples. We find \model makes natural modifications to the sentence and preserves the semantic meanings. 

\begin{table}[htb]
    \centering\small
    \begin{tabular}{p{0.96\columnwidth}}
    \toprule
    \textbf{Original (prediction: Technology):} GERMANTOWN , Md . A Maryland - based \hlt{private lab} that analyzes criminal - case DNA evidence has fired an analyst for allegedly falsifying test data . \\
    \textbf{Adversarial (prediction: Business):} GERMANTOWN , Md . A Maryland - based \hlt{bio testing company} that analyzes criminal - case DNA evidence has fired an analyst for allegedly falsifying test data . \\\midrule
    \textbf{Original (prediction: Sport):} LeBron James scored 25 points , Jeff McInnis added a season - high 24 and the Cleveland Cavaliers won their sixth straight , 100 - 84 over the Charlotte Bobcats on Saturday night \hlt{.}\\
    \textbf{Adversarial (prediction: World):} LeBron James scored 25 points , Jeff McInnis added a season - high 24 and the Cleveland Cavaliers won their sixth straight , 100 - 84 \hlt{Saturday} over the \hlt{visiting} Charlotte Bobcats on Saturday night \hlt{..}\\\midrule
    \textbf{Original (prediction: Negative):} don ' t be fooled by \hlt{the} impressive cast list - eye see you is pure junk .\\
    \textbf{Adversarial (prediction: Positive):} don ' t be fooled by \hlt{this} impressive cast list - eye see you is pure junk .\\\midrule
    \textbf{Original (prediction: Ask for description):} What is die - casting ?\\
    \textbf{Adversarial (prediction: Ask for entity):} What is \hlt{the technique of} die - casting ?\\\midrule
    \textbf{Original (prediction: Toxic)} go back under \hlt{your} rock u irrelevant party puppet\\
    \textbf{Adversarial (prediction: Harmless)} go back under \hlt{the} rock u irrelevant party puppet\\\bottomrule
    \end{tabular}
    \caption{A few adversarial examples generated by \model with the perturbation in red.}\label{tab:case}
    \vspace{-1em}
\end{table}
\begin{table*}[tb]
\small\centering
\begin{tabular}{rcccccccccccccc}
\toprule
           & \multicolumn{3}{c}{\textbf{AG}}       & \multicolumn{3}{c}{\textbf{MR}}       & \multicolumn{3}{c}{\textbf{TREC}}     & \multicolumn{3}{c}{\textbf{HATE}}     & \multicolumn{2}{c}{\textbf{FAKE}} \\\cmidrule(lr){2-4}\cmidrule(lr){5-7}\cmidrule(lr){8-10}\cmidrule(lr){11-13}\cmidrule(lr){14-15}
           & \textbf{S.} & \textbf{F.} & \textbf{M.} & \textbf{S.} & \textbf{F.} & \textbf{M.} & \textbf{S.} & \textbf{F.} & \textbf{M.} & \textbf{S.} & \textbf{F.} & \textbf{M.} & \textbf{S.} & \textbf{F.}    \\\midrule
TextFooler & 3.93       & 3.58    & 0.90  & 3.3        & 3.49    & \textbf{0.92}  & 3.25       & 2.88    & 0.88  & 3.76       & 3.61    & 0.46  & 3.58         & 3.58      \\
CLARE      & 3.75       & 3.65    & 0.93  & 2.44       & 3.33    & 0.74  & 3.00       & 3.00    & 0.75  & \textbf{3.89}       & \textbf{4.41}    & \textbf{0.81}  & 3.67         & 3.65      \\
R\&R (Ours)& \textbf{4.12}       & \textbf{3.87}    & \textbf{0.99}  & \textbf{3.48}       & \textbf{3.61}    & 0.85  & \textbf{3.59}       & \textbf{3.14}    & \textbf{0.89}  & 3.59       & 3.94    & 0.76  & \textbf{3.81}         & \textbf{3.87}     \\\bottomrule
\end{tabular}
\caption{Human evaluation. ``S.'', ``F.'' and ``M.'' represents the similarity, fluency and label match annotated by human.}\label{tab:human_sim}
\end{table*}

\subsection{Is \model effective in attacking classifiers?}

\begin{figure}[htb]
    \centering
    \includegraphics[width=0.9\columnwidth]{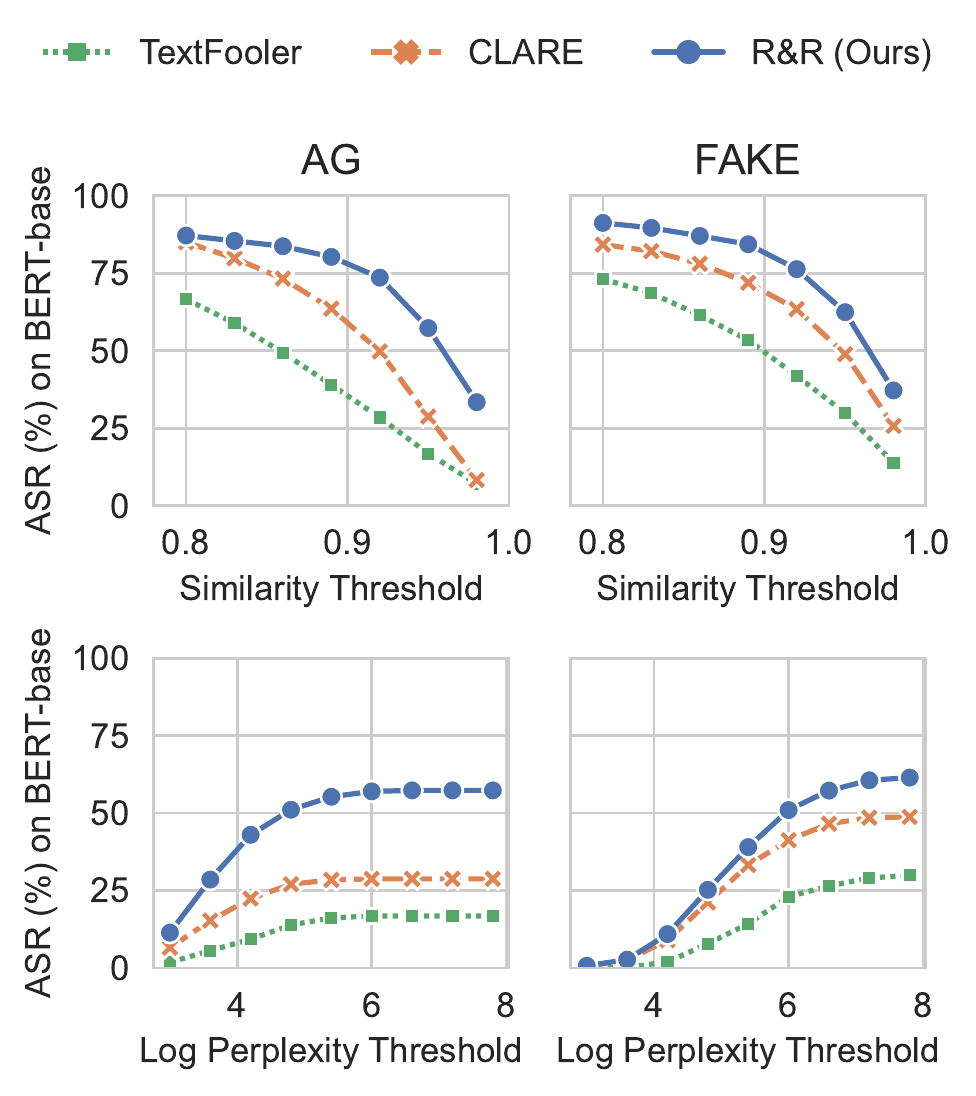}
    \caption{Attack success rate with respect to different similarity and perplexity constraints on BERT classifier. When evaluating different similarity thresholds, we do not set thresholds on perplexity. When evaluating perplexity thresholds, we fix the similarity threshold to 0.95. See Figure~\ref{fig:asr2} in Appendix for other datasets and classifiers.}
    \label{fig:asr}
\end{figure}

Table~\ref{tab:auto_metrics} shows the ASR of \model and baseline methods (with a rigorous 0.95 threshold on similarity). \model achieves the best ASR on all datasets and across all classifiers. The average improvement compared with the CLARE baseline is +16.2\%, +12.8\%, +14.0\% on BERT-base, RoBERTa-large and FastText classifiers respectively. This means that with the same rigorous similarity threshold, \model is capable of finding more adversarial examples, i.e. for some text, \model can find adversarial examples with a similarity higher than 0.95 while baseline methods cannot.
We further measure whether \model can outperform baselines with less rigorous similarity thresholds. On Figure~\ref{fig:asr}, we set different thresholds and show the corresponding ASR. We observe that the curves of \model are above the baseline curves in most cases, showing that \model outperforms baselines on most threshold settings. It means \model can achieve a higher ASR with various different similarity thresholds. 

\subsection{Does \model generate semantically similar and fluent adversarial sentences?}

Table~\ref{tab:auto_metrics} shows the USE similarity metric and log perplexity fluency metric (with a rigorous 0.95 threshold on similarity). Since we already apply a high threshold to ensure the adversarial examples are similar to the original sentences, the similarity metrics do not show significant differences. On AG, MR, TREC and FAKE datasets and 3 classifiers (a total of 12 settings), \model outperforms baseline methods in 9 cases. This shows \model keeps sentence fluency as high as baseline methods do, and does not sacrifice sentence fluency for higher ASR. The only failure case is on the HATE dataset, where TextFooler outperforms \model in perplexity. Further investigation shows that it is because of the perplexity of the original sentence. If the original sentence has high perplexity, the corresponding adversarial sentence is likely to have high perplexity. It is possible that the original sentences that \model succeeds on have higher perplexity than those successfully attacked by TextFooler. Therefore, we compute the average log perplexity for original sentences that are successfully attacked, and find that it is 3.24 for TextFooler and 3.94 for \model. So TextFooler achieves low perplexity because it succeeds on original sentences with low perplexity while failing on those with higher perplexity.

USE similarity and log perplexity are proxy measures. To verify them, human annotations are needed. Table~\ref{tab:human_sim} shows the human evaluation results. \model outperforms baselines on similarity and fluency on 4 datasets. This shows that by optimizing the critique score, \model improves the similarity and fluency of adversarial sentences. Our method fails on the HATE dataset despite good automatic metrics. We hypothesize that this dataset collected from Twitter is more noisy than the others, causing the malfunction of automatic similarity and fluency metrics.


\subsection{Do adversarial sentences preserve the original labels?}
Preserving the original label is critical for an adversarial sentence to be legitimate. Table~\ref{tab:human_sim} also shows the human evaluation on label match. At least 76\% of adversarial examples generated by \model preserves the original label thus being legitimate. We also find that the label match  is task dependent. Preserving original labels on AG dataset is easier than others, while the HATE dataset is the most challenging one. 

\subsection{How does each component in \model contribute to the good performance?}
We conduct ablation study on AG and FAKE datasets to understand the contribution of stochastic decision function, and periodic rollback.

\begin{figure}[htb]
    \centering
    \includegraphics[width=\columnwidth]{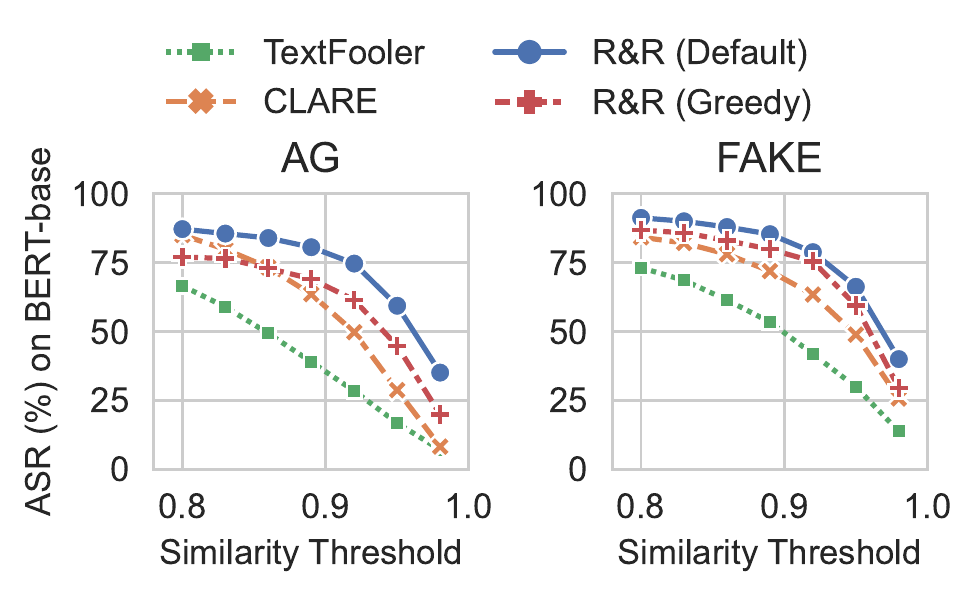}
    \caption{The ASR of \model using different decision settings. ``Greedy'' means using a greedy decision function, which accepts a rewrite only if it has a higher critique score.}
    \label{fig:ab-decision}
\end{figure}

\paragraph{Decision Function}
In the Rewrite stage, we use a stochastic decision function based on the critique score. One alternative can be a deterministic greedy decision function, which accepts a rewrite only if the rewrite increases the critique score.  Figure~\ref{fig:ab-decision} shows the ASR with respect to different similarity thresholds. We find that the stochastic decision function outperforms the greedy one. We interpret the phenomenon as the greedy decision function gets stuck in local maxima, whereas the stochastic one can overcome this issue by accepting a slightly worse rewrite.

\begin{figure}[htb]
    \centering
    \includegraphics[width=\columnwidth]{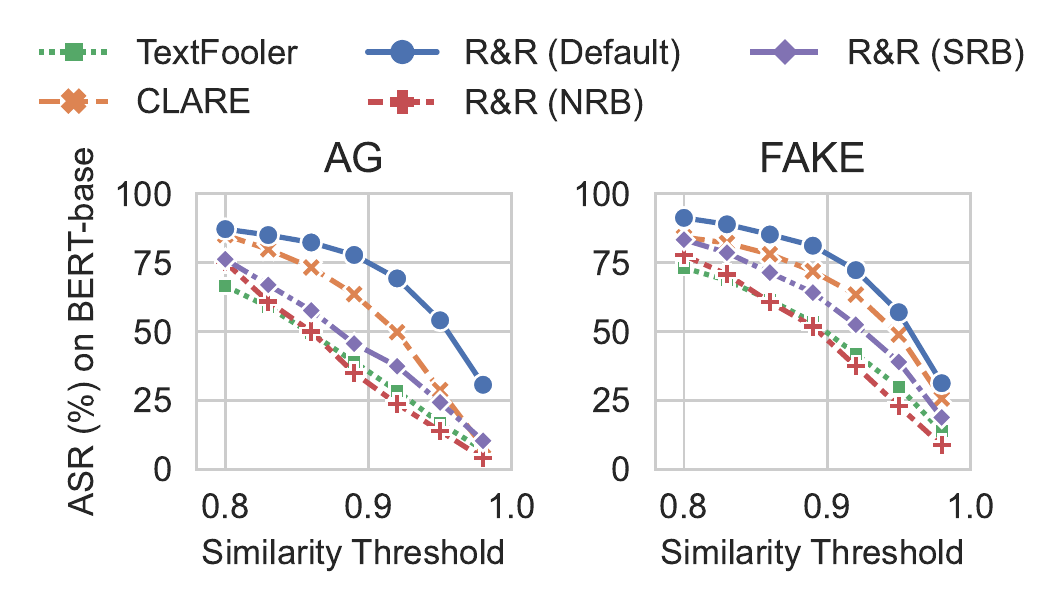}
    \caption{The ASR of \model using different rollback settings. ``NRB'' means no rollback operation and ``SRB'' means single rollback.}
    \label{fig:ab-rollback}
\end{figure}

\paragraph{Rollback} 
We apply rollback periodically during the attack. We compare it with two alternatives: (1) no rollback (NRB) which only uses rewrite to construct the adversarial sentences, and (2) single rollback (SRB) which applies rollback once on the NRB results. 
Figure~\ref{fig:ab-rollback} shows the result. We find that rollback has a significant impact. NRB performs the worst. Without rollback, it is difficult to get  high cosine similarity when many words in the sentence have been changed. Single rollback increases the number of overlapped words, which usually increases the similarity measurement. By periodically applying the rollback, the rollbacked sentence can be further rewritten to improve the similarity and fluency metrics, thus yielding to the best performance.

\begin{figure}[htb]
    \centering
    \includegraphics[width=\columnwidth]{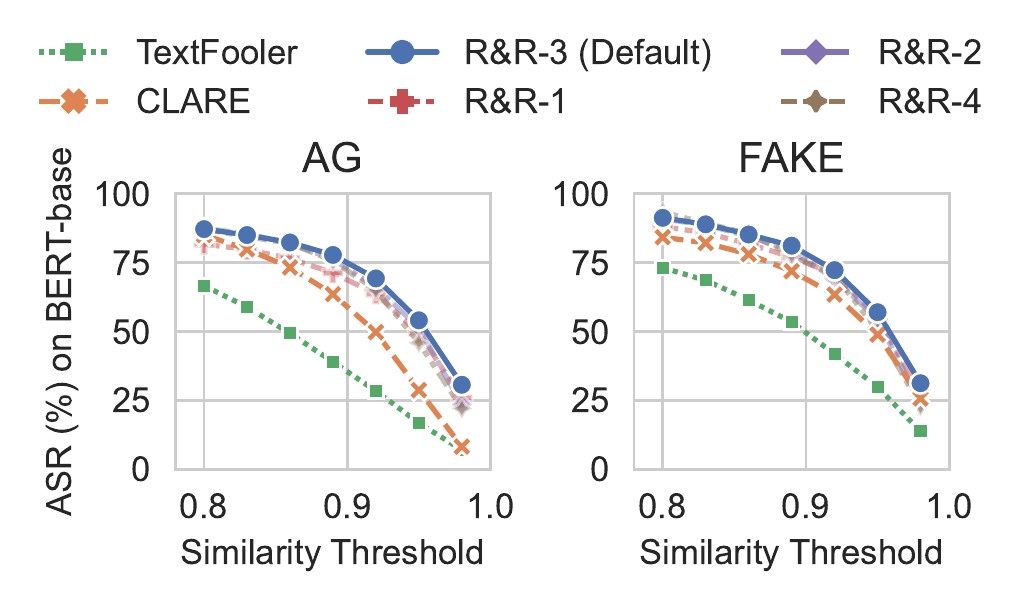}
    \caption{The ASR of \model using different masking span sizes. \model-1 to \model-4 represent the span size of 1 to 4 respectively. We use span size 3 by default.} 
    \label{fig:ab-window}
\end{figure}

\paragraph{Multiple-Word Masking}
In the Rewrite stage, we mask a span of multiple words in each iteration. Intuitively, when using a smaller span size, the masked words are easier to predict. The proposal distribution will assign high probability to the original words at masked positions. Therefore, the candidate sentences are likely to be identical to the original sentence, thus limiting the number of perturbations explored. When the span is large, predicting words becomes more difficult. Thus, we can sample different candidate sentences. But it is more likely to construct dissimilar or influential sentences. 
We vary the span size from 1, 2, 3, to 4 and show the results on Figure~\ref{fig:ab-window}. We find that using span size 3 yields the best performance over most similarity thresholds.  

\subsubsection{How do existing defense methods work against \model?}
We further explore the defense against this attack: 
\begin{itemize}[topsep=1pt, partopsep=1pt, leftmargin=12pt, itemsep=-2pt]
    \item Adversarial attack methods sometimes introduce outlier words to trigger misclassification. Therefore we follow~\citet{qi2020onion} and apply a perplexity-based filtering to eliminate outlier words in sentences. We generate adversarial sentences on vanilla classifiers, then apply the filtering. 
    \item  SHIELD~\citep{le2022shield} is a recently proposed algorithm that modifies the last layer of a neural network to defend against adversarial attack. We apply this method to classifiers and attack the robust classifier. 
\end{itemize}

\begin{table}[htb]
\small\centering
\begingroup
\setlength{\tabcolsep}{4pt}

\begin{tabular}{rrrrr}
\toprule
\multicolumn{1}{l}{} & \multicolumn{2}{c}{\textbf{AG}}     & \multicolumn{2}{c}{\textbf{FAKE}}   \\\cmidrule(lr){2-3}\cmidrule(lr){4-5}
                     & \textbf{+Filter} & \textbf{+SHIELD} & \textbf{+Filter} & \textbf{+SHIELD} \\\midrule
TextFooler           & 6.2      & 8.2      & 13.8     & 16.7     \\
CLARE                & 5.6      & 18.2     & 19.0     & 51.1     \\
\model(ours)          & \textbf{22.3}     &   \textbf{30.6}   & \textbf{23.1}    & \textbf{59.4}  \\\bottomrule
\end{tabular}
\endgroup
\caption{The ASR of attack methods when applying the perplexity-based filtering (Filter) and the SHIELD defense on the BERT classifier.}\label{tab:defense}
\end{table}

Table~\ref{tab:defense} shows the ASR of attack methods with a defense applied. We show that existing defense methods cannot effectively defend against \model. It still outperforms baselines in ASR by large margin.

\section{Related Work}
Several recent works proposed word-level adversarial attacks on text classifiers. This type of attack misleads the classifier's predictions by perturbing the words in the input sentence. TextFooler~\cite{Jin2019IsBR} shows the adversarial vulnerability of the state-of-the-art text classifiers. It uses heuristics to replace words with synonyms to  mislead the classifier effectively. It relies on several pre-trained models, such as word embeddings~\cite{mrkvsic2016counter}, part-of-speech tagger, and Universal Sentence Encoder~\citep{cer2018universal} to perturb the sentence without changing its meaning. However, simple synonym substitution without considering the context results in unnatural sentences. Several works~\cite{garg2020bae, li2020bert,li2021CLARE} address this issue by using masked language models such as BERT~\cite{Devlin2019BERTPO} to propose more natural word substitutions. Our method also belongs to this category. But \model does not maximize the efficacy of each perturbation, instead it allows exploring combinations of perturbations to generate  adversarial examples with high similarity with the original sentence. 
Besides word-level attacks \citep{zang2019word,ren-etal-2019-generating}, there are also character-level attacks which introduce typos to trigger misclassification \citep{papernot2016crafting, liang2017deep, samanta2018generating}, and sentence-level attacks which attack a classifier by altering the sentence structure~\citep{iyyer2018adversarial}. \citet{zhang2020adversarial} gives a comprehensive survey on such  attack methods.  Other work on robustness to adversarial attacks in NLP includes robustness of the machine translation models \citep{cheng2019robust}, robustness in domain adaptation \citep{oren2019distributionally},  adversarial examples generated by reinforcement learning \citep{wong2017dancin, vijayaraghavan2019generating}, and certified robustness \citep{jia2019certified}. Adversarial attack libraries \citep{morris2020textattack,zeng2021openattack} are also developed to help future research.

\section{Conclusion}\label{sec:con}
In this paper, we formulate the textual adversarial attack as a multi-objective optimization problem. We use a critique score to synthesize the similarity, fluency, and misclassification objectives, and propose \model that optimizes the critique score to generate high-quality adversarial examples. We conduct extensive experiments. Both automatic and human evaluation show that the proposed method succeeds in optimizing the automatic similarity and fluency metrics to generate adversarial examples of higher quality than previous methods.


\section*{Ethical Considerations}
In this paper, we propose \model to generate adversarial sentences. Like all other adversarial attack methods, this method could be abused by malicious users to attack NLP systems and obtain illegitimate benefits. However, it is still necessary for the research community to develop methods to exploit all vulnerabilities of a classifier based on which more robust classifiers can be developed. 

\section*{Acknowledgments}
Alfredo Cuesta-Infante has been funded by the Spanish Government research project MICINN PID2021-128362OB-I00.

\section*{History of this Paper}
ArXiv:2010.11869v1 is an earlier version of this work. It introduces the \textit{proposal distribution} and proposes to attack text classifiers by rewriting a sentence. 

ArXiv:2104.08453v1 improves the framework by proposing the \textit{critique score} and a stochastic \textit{decision function}. It also conducts human evaluation to verify the quality of adversarial sentences. 

ArXiv:2104.08453v2 fixes some typos.

ArXiv:2104.08453v3 is the current version and is accepted to \textit{Findings of AACL-IJCNLP2022}. In this version, we introduce the rollback step because it can improve the quality of adversarial sentences. We also add a stronger baseline -- CLARE. 

\bibliographystyle{acl_natbib}
\bibliography{aaai22}

\clearpage
\appendix
\section{Relation to Metropolis-Hastings Sampling}\label{app:mhs}

Metropolis-Hastings sampling (MHS)~\citep{hastings1970monte} is a Markov-chain Monte Carlo (MCMC) for generating independent unbiased samples from a distribution. Assume we have a target distribution of sentences $p_\text{target}(\bu|\bx, y)$ such that legitimate adversarial sentences of $\bx$ have high probability, while other sentences (could be a meaningless sequence of words) have low probability, we may attempt to solve the adversarial attack problem by MHS. Because we are likely to get an adversarial sentence of $\bx$ by drawing samples from $p_\text{target}(\bu|\bx,y)$. To apply MHS, we need to choose a transition probability $p_\text{transition}(\hat\bu|\bu,\bx,y)$ that defines the probability to transit from one sentence to the next sentence in the MCMC. Then the MHS has following steps:
\begin{enumerate}
    \item Start with $\bu=\bx$.
    \item Get a candidate $\hat\bu\sim p_\text{transition}(\hat\bu|\bu,\bx,y)$.
    \item Compute 
    \begin{equation}
         \alpha = \frac{p_\text{target}(\hat\bu|\bx,y) p_\text{transition}(\bu|\hat \bu,\bx,y)}{p_\text{target}(\bu|\bx,y) p_\text{transition}(\hat\bu|\bu,\bx,y)}.  \label{eq:alpha2}
    \end{equation}
    \item With probability $\min(\alpha, 1)$, use $\hat\bu$ as new $\bu$ and go to step 2; otherwise use the previous $\bu$ and go to step 2.
    \item After sufficient iterations, $\bu$ is a sample drawn from $p_\text{target}(\bu|\bx,y)$. Note that MHS needs a lot of iterations considering the huge space of all sentences. 
\end{enumerate}

The rewrite step in \model is similar to MHS, if we consider $\exp[C(\bu)]$ as the unnormalized target distribution\footnote{We apply the exponential function to make sure the probability mass is positive.} and $p_\text{proposal}(\cdot)$ as the transition probability. The definition of $\alpha$ in Eq.~(\ref{eq:alpha}) and Eq.~(\ref{eq:alpha2}) is one significant difference, where \model only uses target distribution and omits the transition probability. We find omitting it can make the sampling bias towards sentences with higher probability in target distribution (i.e., sentences with higher critique score), which benefits the adversarial attack efficacy.

\begin{figure*}[htb]
    \centering
    \textbf{Similarity}\\
    \includegraphics[width=0.9\textwidth]{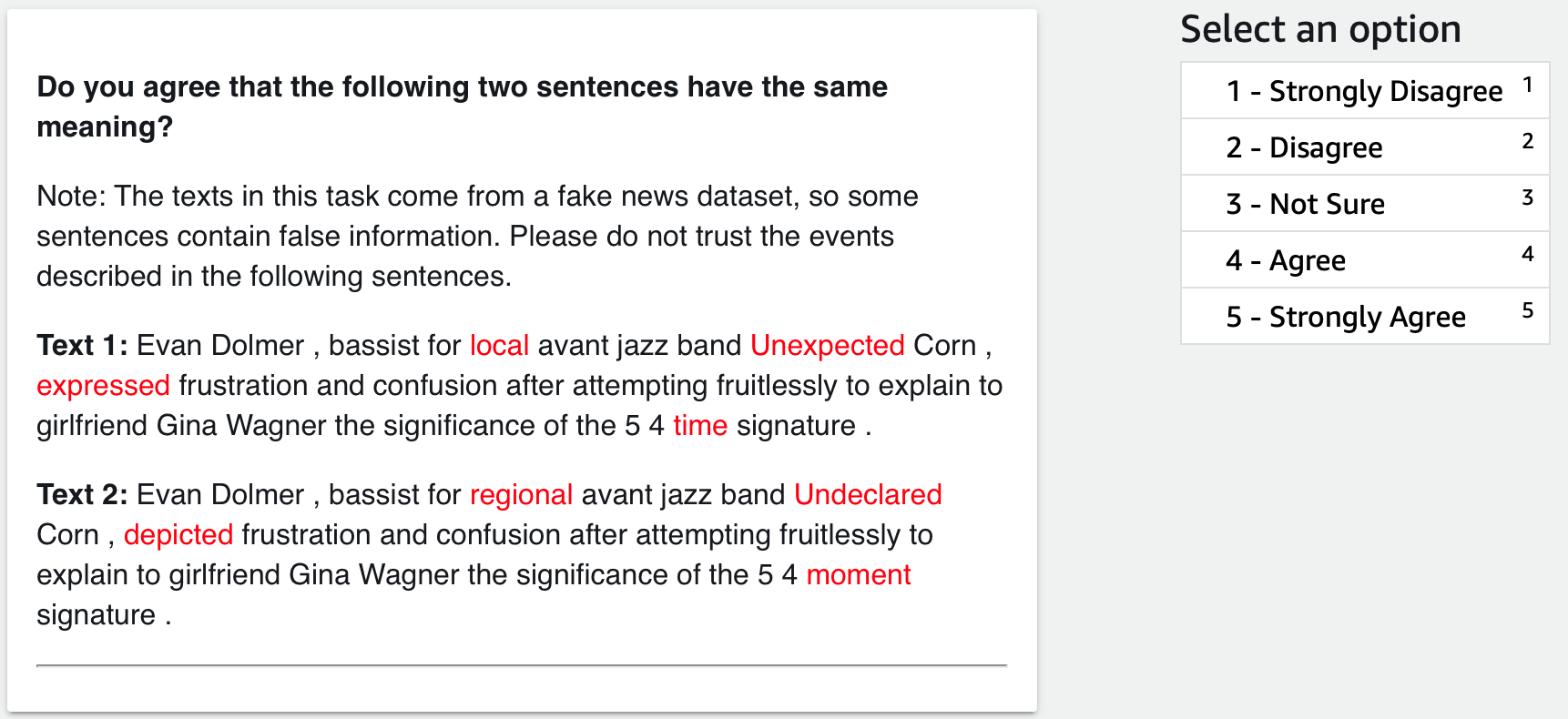}
    
    \textbf{Fluency}\\
    \includegraphics[width=0.9\textwidth]{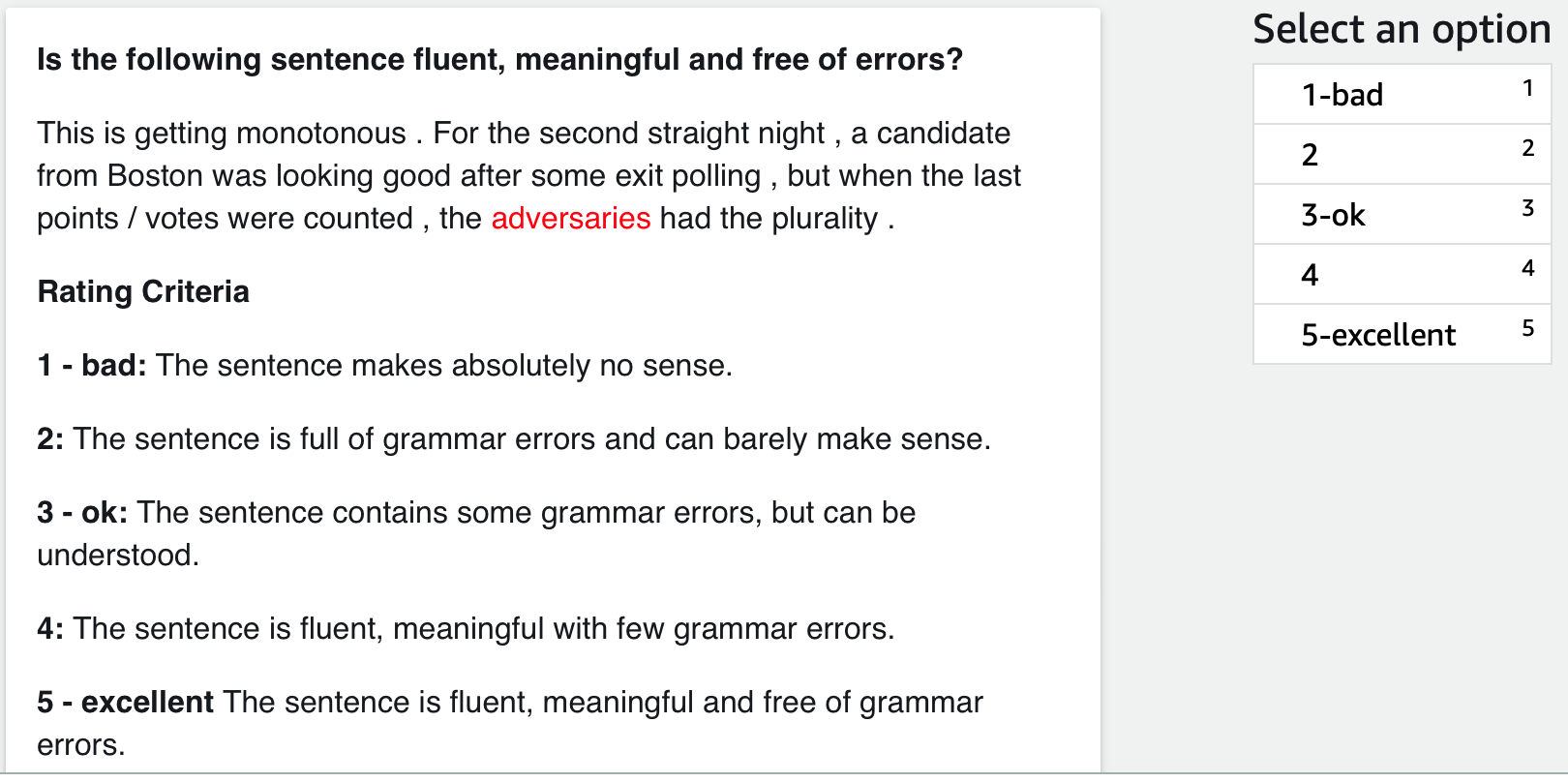}
    
    \textbf{Label Match}\\
    \includegraphics[width=0.9\textwidth]{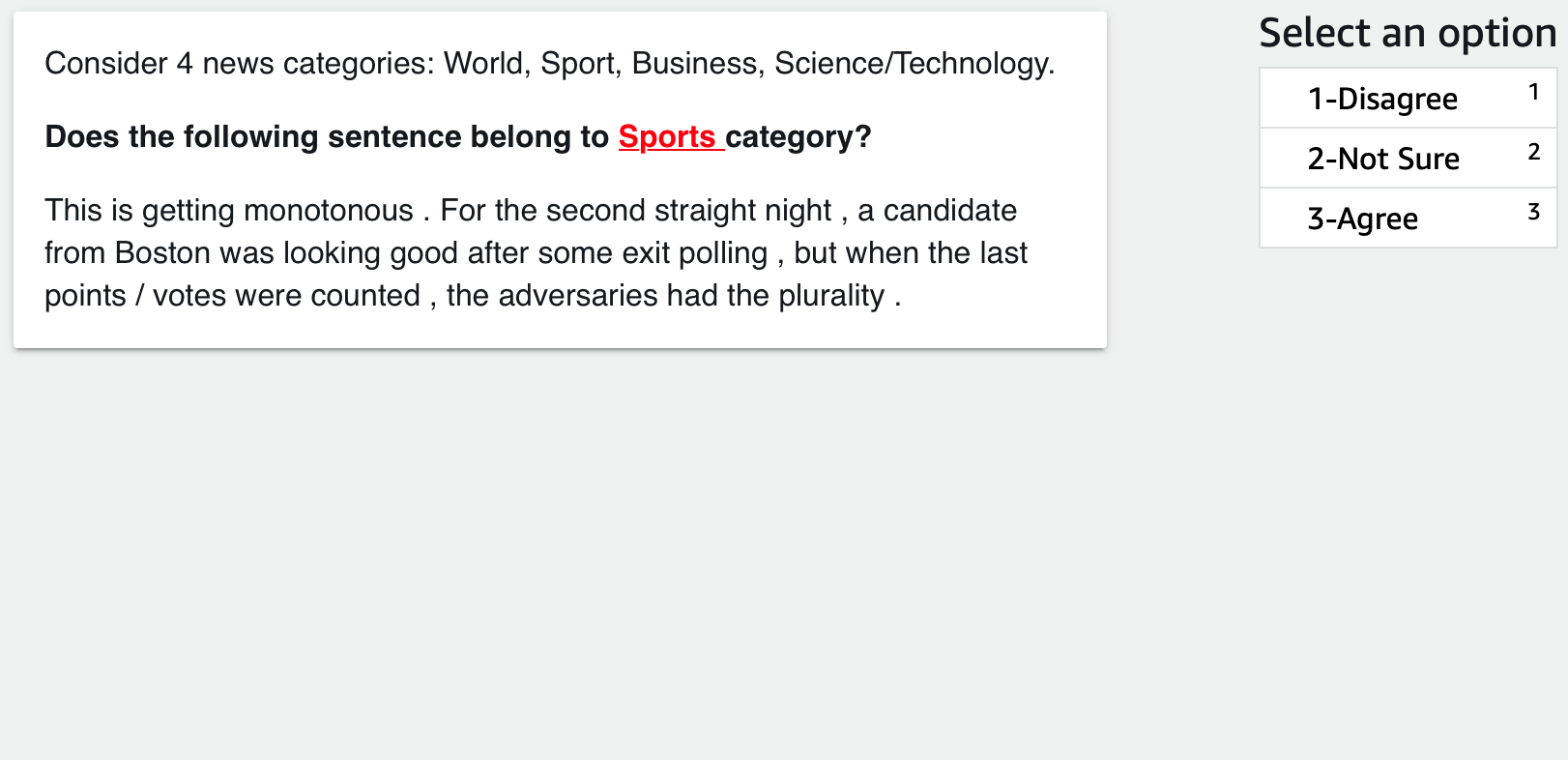}
    \caption{The screenshots of MTurk tasks.}
    \label{fig:mturk_screen}
\end{figure*}

\begin{figure*}[tb]
    \centering
    Classifier: BERT
    \includegraphics[width=\textwidth]{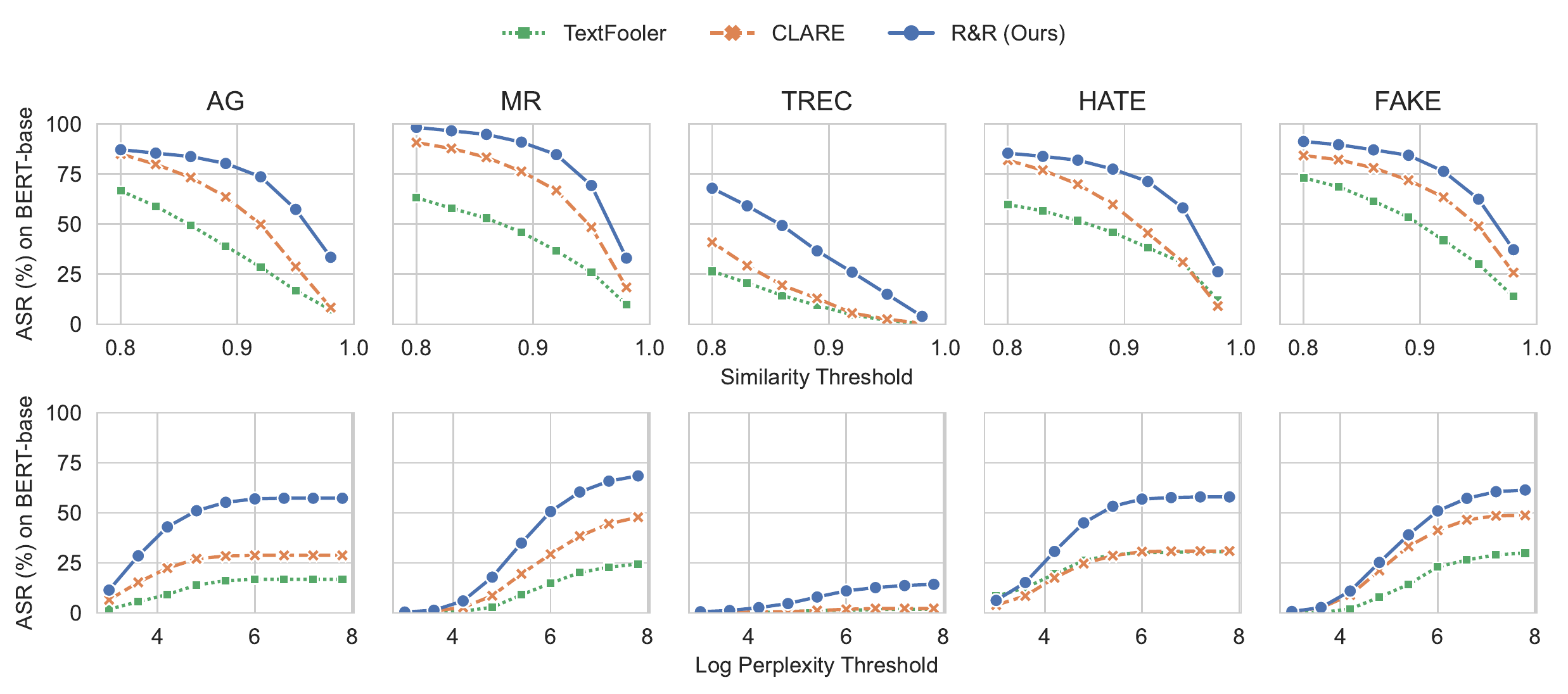}
    Classifier: RoBERTa
    \includegraphics[width=\textwidth]{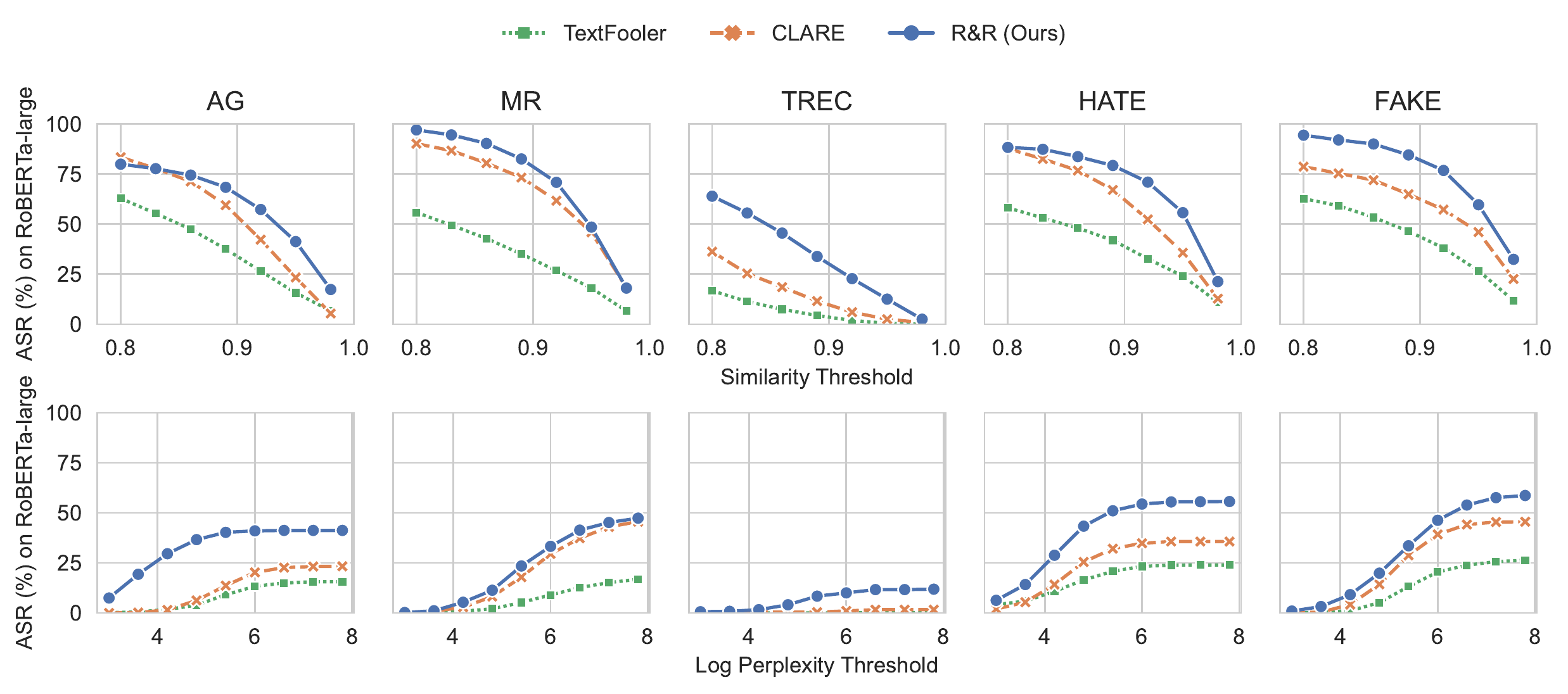}
    Classifier: FastText
    \includegraphics[width=\textwidth]{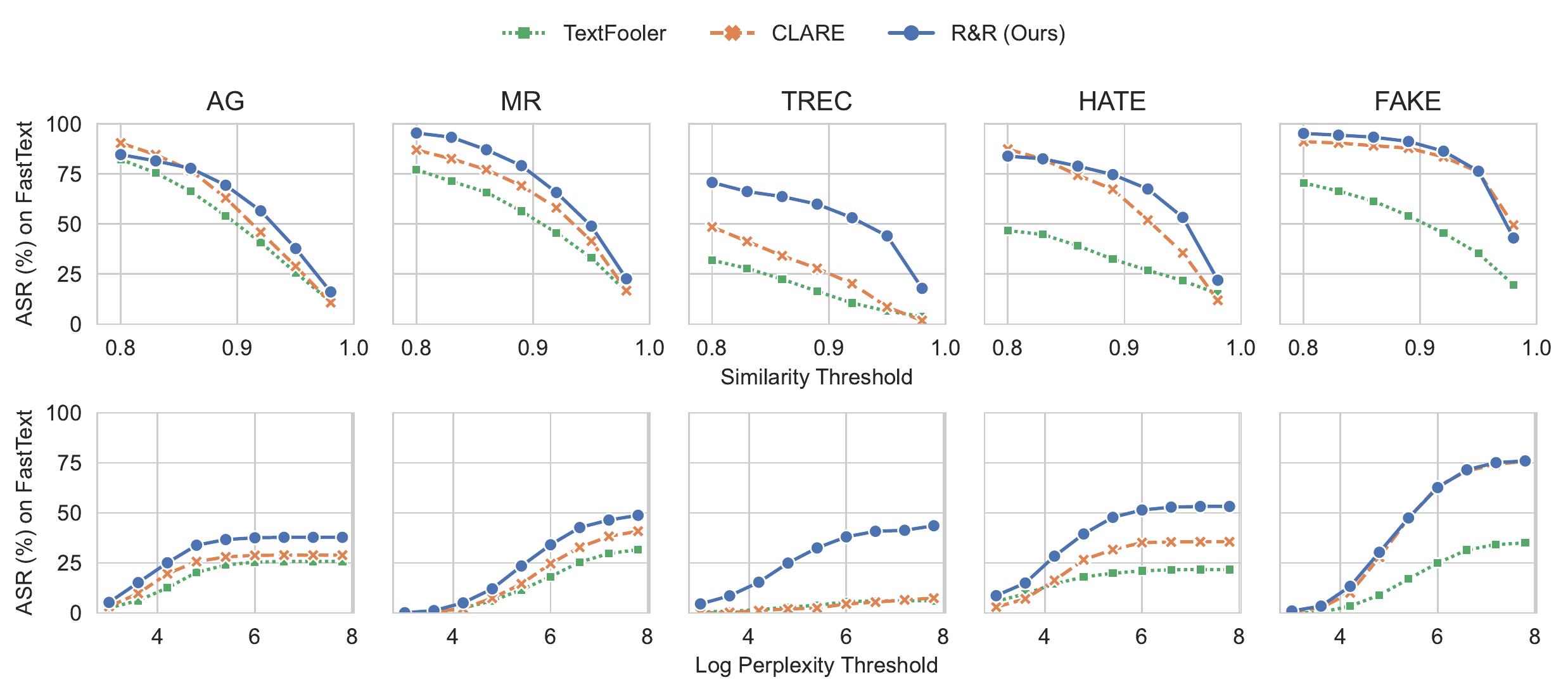}
    \caption{Attack success rate with respect to different similarity and perplexity constraints. When evaluating different similarity thresholds, we do not set thresholds on perplexity. When evaluating perplexity thresholds, we fix the similarity threshold to 0.95. }
    \label{fig:asr2}
\end{figure*}

\begin{figure*}[tb]
    \centering
     \includegraphics[width=0.9     \textwidth]{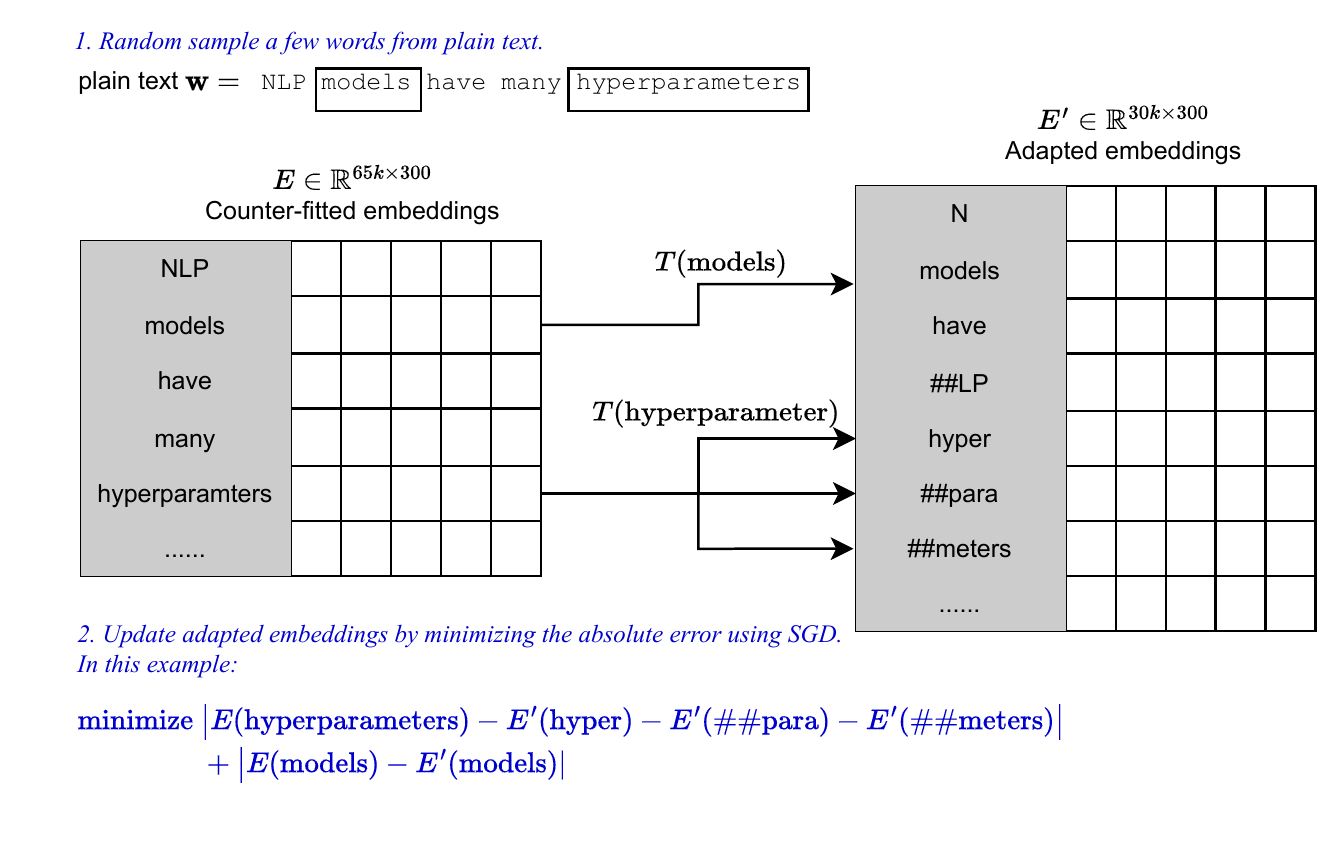}
    \caption{One learning step of vocabulary adaptation algorithm. The plain text has only 5 words in this example, but it has much more words in real datasets. We illustrate by sampling only 2 words from plain text, while we sample 5000 words in practice. }
    \label{fig:adapt}
\end{figure*}

\end{document}